\pdfoutput=1
\documentclass[11pt]{article}

\usepackage[preprint]{acl}

\usepackage{times}
\usepackage{booktabs}
\usepackage{amsmath}
\usepackage{latexsym}
\usepackage{graphicx}
\usepackage{multirow}
\usepackage{tabularx}
\usepackage[table]{xcolor}

\usepackage{array}
\usepackage[most]{tcolorbox}
\usepackage{xcolor}
\usepackage{listings}
\lstdefinelanguage{json}{
  basicstyle=\ttfamily\small,
  numbers=none,
  stepnumber=1,
  numbersep=8pt,
  showstringspaces=false,
  breaklines=true,
  frame=single,
  backgroundcolor=\color{black!2},
  stringstyle=\color{green!50!black},
  keywordstyle=\color{blue!70},
  commentstyle=\color{black!60},
  morekeywords={true,false,null},
}
 \usepackage{amssymb}
 \definecolor{c1}{RGB}{239,118,122}
\definecolor{c2}{RGB}{69,105,144}
\definecolor{c3}{RGB}{72,192,170}
\definecolor{c4}{RGB}{179,149,189}
\usepackage[T1]{fontenc}
\usepackage[utf8]{inputenc}

\usepackage{microtype}

\usepackage{inconsolata}

\usepackage{graphicx}
\usepackage{amsmath,amsfonts,bm}

\def\eqref#1{equation~\ref{#1}}
\def\1{\bm{1}}

\DeclareMathAlphabet{\mathsfit}{\encodingdefault}{\sfdefault}{m}{sl}
\SetMathAlphabet{\mathsfit}{bold}{\encodingdefault}{\sfdefault}{bx}{n}

\title{MemoBrain: Executive Memory as an Agentic Brain for Reasoning
}

\author{Hongjin Qian$^{1}$,  Zhao Cao$^{2}$, Zheng Liu$^1$\thanks{Corresponding author.}\\
        $^1$ Beijing Academy of Artificial Intelligence \\ 
        $^2$  Gaoling School of Artificial Intelligence, Renmin University of China\\
        \texttt{\{chienqhj,zhengliu1026\}@gmail.com} \\
}

\begin{document}
\maketitle

\begin{abstract}
Complex reasoning in tool-augmented agent frameworks is inherently long-horizon, causing reasoning traces and transient tool artifacts to accumulate and strain the bounded working context of large language models.
Without explicit memory mechanisms, such accumulation disrupts logical continuity and undermines task alignment. This positions memory not as an auxiliary efficiency concern, but as a core component for sustaining coherent, goal-directed reasoning over long horizons.

We propose \textbf{MemoBrain}, an \emph{executive memory} model for tool-augmented agents that constructs a dependency-aware memory over reasoning steps, capturing salient intermediate states and their logical relations.
Operating \emph{as a co-pilot} alongside the reasoning agent, MemoBrain organizes reasoning progress without blocking execution and actively manages the working context.
Specifically, it prunes invalid steps, folds completed sub-trajectories, and preserves a compact, high-salience reasoning backbone under a fixed context budget.
Together, these mechanisms enable explicit cognitive control over reasoning trajectories rather than passive context accumulation.

We evaluate MemoBrain on challenging long-horizon benchmarks, including GAIA, WebWalker, and BrowseComp-Plus, demonstrating consistent improvements over strong baselines. Our codes are in  \href{https://github.com/qhjqhj00/MemoBrain}{\textit{this repository}}.
\end{abstract}

\begin{figure*}[t]
    \centering
    \includegraphics[width=\linewidth]{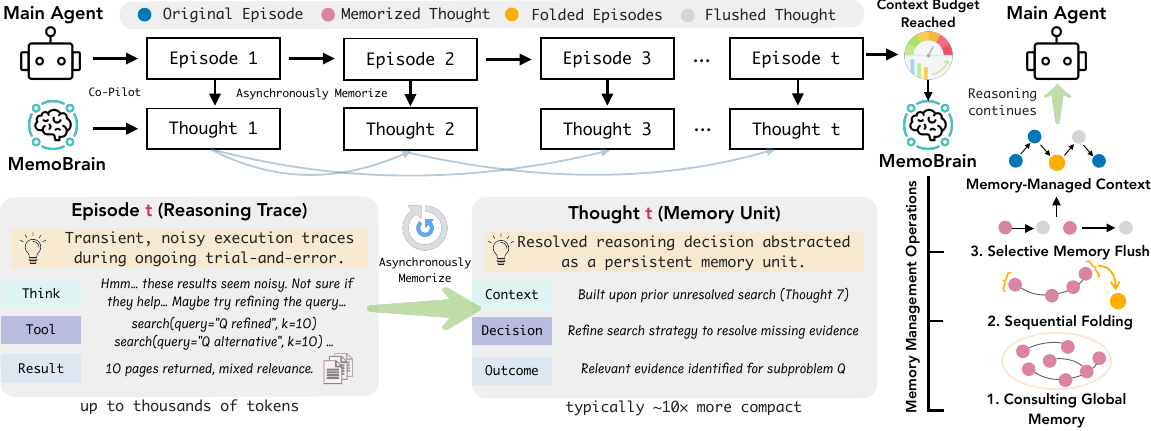}
    \caption{MemoBrain asynchronously abstracts transient reasoning sessions into compact, dependency-aware memory units and maintains a global reasoning memory.
Upon reaching the memory budget, it constructs a memory-managed context by folding or flushing past information, enabling coherent long-horizon reasoning.}
    \label{fig:model}
    \vspace{-10pt}
\end{figure*}

\section{Introduction}
Large language models (LLMs) have advanced rapidly in recent years, enabling increasingly complex applications in reasoning, decision making, and information seeking~\citep{zhao2024surveylargelanguagemodels,agentsurvey}.
Tool-augmented agent frameworks further extend these capabilities by interleaving reasoning with calls to external tools such as search engines and code executors~\citep{agentrlsurvey,qian2025toolrlrewardtoollearning}.
However, long-horizon tool-integrated interactions inevitably accumulate contextual information, including intermediate reasoning traces and tool-generated artifacts~\citep{agentfold}.
As context grows, the bounded working context of LLMs becomes strained, allowing transient information to dominate and distract from task-relevant evidence, leading to cognitive overload and degraded task alignment~\citep{hu2025memoryageaiagents}.

As a result, introducing explicit memory mechanisms has become widely recognized as a critical factor in scaling agent-based frameworks to complex, long-horizon tasks~\citep{agentmemory}.
Existing agent memory mainly takes two forms.
\emph{Cross-task memory} focuses on persistence across tasks, accumulating experience, preferences, and user-specific knowledge to support personalization, adaptation, and learning from past interactions~\citep{selfevolve}.
\emph{Long-term memory}, in contrast, is designed to retain and organize information over long interactions, such as long-term dialogue or large collections of documents, and is commonly instantiated through summarization, parametric storage, or graph-based representations~\citep{xu2025amemagenticmemoryllm,gam,yu2025memagentreshapinglongcontextllm,kang2025memoryosaiagent}.
While effective at preserving and accessing prior information, both forms are not natively adaptive to complex reasoning settings, where memory must actively track reasoning progress, capture dependencies among intermediate conclusions, and manage information flow to sustain coherence and task alignment.

In this work, we focus on a distinct form of memory, which we term \emph{executive memory}.
\textbf{Executive memory is a task-specific, in-process memory mechanism dedicated to complex reasoning in tool-augmented agent systems, initialized from scratch for each task and evolving online alongside the reasoning trajectory}.
Unlike cross-task memory, it does not persist beyond task completion or encode persistent priors.
Unlike long-term memory, it is not merely a conditioning context that is incrementally maintained, but an actively managed representation that adapts to the evolving structure of the reasoning process.
From an engineering perspective, executive memory functions as an explicit \emph{context control mechanism}, delivering task-relevant information to the reasoning agent in a just-in-time manner under a bounded context budget.
From a cognitive perspective, it serves an executive role by modeling logical dependencies among reasoning steps, maintaining a global and hierarchical view of task progress, and guiding subsequent actions to preserve coherent, goal-directed reasoning.

A growing body of recent work has highlighted the importance of managing large and evolving contexts in complex reasoning tasks.
Representative approaches reduce context overhead by abstracting or compressing intermediate reasoning steps~\citep{resum,agentfold}, while others decompose tasks into auxiliary subtasks whose outcomes are summarized and reintegrated~\citep{li2025webthinker,deepagent}.
These methods have shown strong empirical gains on long-horizon reasoning benchmarks, underscoring the practical value of memory from an engineering perspective.
However, most existing approaches frame memory as an efficiency-oriented mechanism for context reduction, leaving it largely passive with respect to global task alignment, trajectory-level control, and cognitive modeling of the reasoning process~\citep{hu2025memoryageaiagents}.

These observations suggest that sustaining task-aligned reasoning under bounded context budget requires explicit control over information flow.
Accordingly, we treat memory as a framework-level component that collaborates with the reasoning agent, rather than an internal byproduct of a single model.
Based on this perspective, we propose \textbf{MemoBrain, an executive memory model that operates in a copilot-like manner alongside the reasoning agent}.
MemoBrain maintains a global memory over the full reasoning trajectory and exerts executive control over the working context, delivering just-in-time, task-relevant information while suppressing unnecessary cognitive load.
By controlling what information is exposed at each step, MemoBrain enables goal-directed reasoning over long horizons under a bounded context budget.

As illustrated in Figure~\ref{fig:model}, MemoBrain organizes memory around two complementary processes: memory construction and memory management.
During memory construction, completed reasoning episodes are incrementally abstracted from transient execution sessions into compact memory units, referred to as \emph{thoughts}.
Each thought encodes the subproblem addressed, the information or tools used, and the resulting outcome, and is linked to prior thoughts through explicit dependency relations.
This process yields a structured, dependency-aware trajectory memory that preserves the logical backbone of long-horizon reasoning while discarding short-lived execution artifacts.

Building on this global trajectory memory, MemoBrain performs active context management for the reasoning agent.
Under a fixed context budget, it selectively folds completed sub-trajectories, flushes low-utility memory elements, and prioritizes high-salience reasoning backbones.
Together, these operations ensure that the working context remains compact yet logically sufficient, allowing the agent to sustain coherent and task-aligned reasoning as the reasoning horizon grows.

To optimize memory construction and memory management for their distinct functional demands, MemoBrain adopts a task-specific optimization strategy that aligns learning objectives with downstream reasoning behavior, enabling memory to function as an active and adaptive component for long-horizon reasoning.
We evaluate MemoBrain across model scales ranging from 4B to 14B by integrating it into tool-augmented agents with strong long-horizon reasoning capabilities.
Across challenging benchmarks including BrowseComp-Plus~\citep{bcp}, GAIA~\citep{mialon2023gaia}, and WebWalker~\citep{wu2025webwalker}, MemoBrain consistently improves performance on complex information-seeking and reasoning tasks under bounded context budget.

Our contributions are threefold:
(1) We introduce \emph{executive memory} as a new memory paradigm dedicated to complex reasoning in tool-augmented agent frameworks, formalizing memory as an explicit control mechanism for context management and cognitive modeling under a bounded context budget.
(2) We propose \emph{MemoBrain}, a standalone executive memory model that operates as a co-pilot alongside the reasoning agent, constructing dependency-aware in-task memories and actively manages the working context through explicit memory operations.
(3) We empirically demonstrate that integrating executive memory into diverse tool-augmented agent frameworks consistently improves performance across multiple model scales and challenging benchmarks, highlighting its generality and effectiveness across agent ecosystems.

\section{Related Work}

Memory has emerged as a foundational mechanism for scaling large language models~\citep{packer2023memgpt, agentmemory, mem0}, particularly in settings involving long contexts and complex information access~\citep{hu2025memoryageaiagents}.
A large body of work focuses on \textbf{agentic long-term memory}, where models are equipped with mechanisms to retain and organize information across extended interactions in agent frameworks~\citep{agentrlsurvey}.
In these settings, memory is typically constructed over long documents, retrieved passage collections, or multi-turn dialogue histories, and is realized through structured abstractions such as graph-based organization, key–value compression, or hierarchical storage~\citep{memorybank,longtermmemory,mem0,edge2024localglobalgraphrag,qian2024memoragmovingnextgenrag,kang2025memoryosaiagent,gam}.
Such memory mechanisms are primarily designed to support persistent knowledge accumulation and reuse, serving as long-term priors for tasks including conversational agents, long-context question answering, and interactive assistants~\citep{li2025memosoperatingmemoryaugmentedgeneration,yan2025memoryr1enhancinglargelanguage,xu2025amemagenticmemoryllm}.
Accordingly, they are commonly evaluated on benchmarks such as LoCoMo~\citep{locomo}, LongBench~\citep{bai2025longbenchv2deeperunderstanding}, and RULER~\citep{hsieh2024ruler}, where memory is built from provided interaction histories or long contexts and accessed during inference.

In tool-augmented complex reasoning, prior work has highlighted the importance of \textbf{context management}~\citep{mei2025survey}.
Methods such as AgentFold and ContextFolding reduce context overhead by folding or abstracting redundant reasoning steps~\citep{agentfold,sun2025scalinglonghorizonllmagent}, while approaches like DeepAgent delegate subtasks to auxiliary agents and reintegrate only summarized outcomes~\citep{li2025webthinker,deepagent}.
Although effective, these methods focus on the engineering aspect of memory and do not model it as a cognitive control mechanism for global task awareness or reasoning guidance.
In contrast, we formalize memory in tool-augmented reasoning as \textbf{executive memory}, an online mechanism that constructs structured representations of reasoning trajectories and explicitly controls information flow.
By unifying context management with cognitive modeling, executive memory enables coherent long-horizon reasoning beyond purely engineering-level optimizations.

\section{Method}
\subsection{Overview of MemoBrain}
Consider a complex reasoning task $\mathcal{X}$ that requires a sequence of
\emph{reasoning episodes}
$\{x_1, x_2, \ldots, x_T\}$ to collect sufficient evidence and produce a final
answer $\mathcal{Y}$.
As illustrated in Figure~\ref{fig:model}, each reasoning episode $x_t$ constitutes
a self-contained unit of progress, typically involving task reasoning,
interaction with external tools (e.g., web search or code execution),
and the tool feedback, together producing diverse and evolving
execution traces.

\paragraph{Memory Construction.} 
MemoBrain incrementally transforms the sequence of reasoning episodes into a structured
trajectory memory.
Each completed episode $x_t$ is abstracted into a compact memory unit, referred to as a
\emph{thought} $v_t$, capturing its semantic contribution to task progress. As reasoning proceeds, newly generated thoughts are integrated into a global,
dependency-aware memory structure,
\begin{equation}
\mathcal{G}_t = \mathcal{G}_{t-1} \cup \{v_t\},
\end{equation}
yielding an evolving memory graph $\mathcal{G}$ that represents the high-level
reasoning trajectory beyond transient execution details.

\paragraph{Memory Management.}
The reasoning agent operates under a bounded context budget.

When contextual accumulation approaches the budget limit,
MemoBrain activates an executive memory management process to regulate how
the accumulated reasoning structure is maintained and exposed.
At a high level, MemoBrain conditions on the current trajectory memory
$\mathcal{G}_t$ and invokes the memory model $\phi(\cdot)$ to produce a set of executive operations:
\begin{equation}
\mathcal{O}_t = \phi(\mathcal{G}_t),
\label{eq:operations}
\end{equation}
which specify how parts of the trajectory memory should be reorganized.
Applying these operations yields an updated memory state:
\begin{equation}
\mathcal{G}_{t+1} = \mathrm{Apply}(\mathcal{G}_t, \mathcal{O}_t).
\end{equation}
Based on $\mathcal{G}_{t+1}$, MemoBrain reorganizes the working context to retain
a compact yet semantically sufficient representation for subsequent reasoning.

\subsection{Memory Construction}
\label{sec:memory_initialization}

\paragraph{Thought Formation.}
After a reasoning episode completes, MemoBrain abstracts it into a \emph{thought}.
Each reasoning episode $x_t$ is represented as:
\begin{equation}
x_t = (\tau_t, \omega_t),
\end{equation}
where $\tau_t$ denotes transient execution-level information
(e.g., tool calls and raw outputs), and $\omega_t$ denotes the resolved semantic
outcome that advances the task.
Rather than retaining execution details, MemoBrain focuses on capturing the
task-relevant contribution of $\omega_t$.
Conditioned on the current memory state $\mathcal{G}_{t-1}$ and a dedicated
memorization instruct (see Table~\ref{prompt:memorize}), MemoBrain applies:
\begin{equation}
v_t = \phi(x_t, \mathcal{G}_{t-1}),
\end{equation}
where $\phi(\cdot)$ maps the completed episode into an executive-level abstraction,
capturing \emph{what information was used to address which subproblem and with
what outcome}.
Each thought encodes the functional role of the episode within the overall reasoning
trajectory, preserving its contribution to task progress while discarding
short-lived execution artifacts.
The newly formed thought $v_t$ is then integrated into memory by updating the
global state from $\mathcal{G}_{t-1}$ to $\mathcal{G}_t$.
This process runs asynchronously alongside the reasoning agent, enabling
MemoBrain to operate as a copilot that continuously records and structures
reasoning progress without interrupting reasoning execution.

Each newly created thought is initialized as \texttt{active}, indicating its
participation in the current reasoning trajectory.
Thoughts can take one of two activation states, \texttt{active} or \texttt{inactive}, which may be updated during memory management.
This allows completed, superseded, or unproductive reasoning paths to be
deactivated without removing their structural role in the memory graph.

\paragraph{Dependency Modeling.}
To preserve the logical structure of reasoning, MemoBrain models how
new thoughts build upon existing ones:
\begin{equation}
\mathrm{Dep}(v_t) \subseteq \{v_1, \ldots, v_{t-1}\},
\end{equation}
representing the earlier conclusions on which the current thought depends.
These dependency relations encode causal and semantic structure beyond mere
temporal order.

As reasoning proceeds, accumulated thoughts and dependencies form an evolving
directed memory graph
$
\mathcal{G}_t = (\mathcal{V}_t, \mathcal{E}_t),
$
where $\mathcal{V}_t = \mathcal{V}_{t-1} \cup \{v_t\}$ and
$\mathcal{E}_t = \mathcal{E}_{t-1} \cup \{(v_i, v_t) \mid v_i \in \mathrm{Dep}(v_t)\}$.
This graph provides a global, trajectory-level representation of the reasoning
process that extends beyond the agent’s local context window.

\subsection{Memory Management}
\label{sec:memory_management}

Memory management specifies how executive operations are applied to the
trajectory memory to regulate the persistence and exposure of reasoning
structure under bounded context.

At each management step, MemoBrain produces a set of executive operations
$\mathcal{O}_t$ over the current trajectory memory $\mathcal{G}_t$, instructed by the management prompt (see Table~\ref{tab:recall}).
Each operation in $\mathcal{O}_t$ acts on selected nodes or subgraphs and
rewrites the memory structure accordingly.
In this work, we consider two classes of operations:
\begin{equation}
\mathcal{O}_t \subseteq \{\textsc{Fold}(\cdot),\; \textsc{Flush}(\cdot)\}.
\end{equation}

\paragraph{Sequential Trajectory Folding.}
The \textsc{Fold} operation targets sub-trajectories whose associated subtasks
have reached a decisive resolution.
Formally, MemoBrain identifies a connected subgraph
$\mathcal{T}_{i:j} \subset \mathcal{G}_t$ composed of a sequence of thoughts
$\{v_i, \ldots, v_j\}$ that jointly address the same subproblem.
If the terminal thought $v_j$ encodes a conclusive outcome
(e.g., target information verified, or evidence consistently absent),
then the intermediate reasoning episodes in $\mathcal{T}_{i:j}$ no longer provide
additional informational gain.
In such cases, MemoBrain applies a folding operation:
\begin{equation}
\mathcal{T}_{i:j} \;\Rightarrow\; \bar{v},
\end{equation}
collapsing the entire sub-trajectory into a single summary thought $\bar{v}$
that records the resolved conclusion.
All thoughts in $\mathcal{T}_{i:j}$ are removed from the active reasoning context.
Because the subtask has been conclusively settled, the corresponding reasoning
sessions can be safely eliminated without compromising task-relevant semantics.

\paragraph{Selective Memory Flush.}
The \textsc{Flush} operation targets individual thoughts whose informational
utility has expired with respect to the current reasoning state.
Specifically, MemoBrain identifies thoughts $v_k$ that fall into one of the
following categories:
(i) invalid reasoning steps that failed to yield meaningful feedback,
(ii) exploratory attempts superseded by subsequent refinements, or
(iii) steps whose conclusions are no longer relevant to the active trajectory.

Unlike folding, flush operations do not assume subtask resolution.
Instead of removing such nodes entirely, MemoBrain applies:
\begin{equation}
v_k \;\Rightarrow\; \hat{v}_k,
\end{equation}
replacing $v_k$ with a compact thought $\hat{v}_k$ that preserves high-level
structural information while discarding execution-level detail.
These compact thoughts record the existence and outcome of prior attempts,
preventing redundant exploration while avoiding unnecessary contextual load.

\paragraph{Context Reorganization.}
Following executive memory transformation, MemoBrain reorganizes the working
context by projecting the updated trajectory memory $\mathcal{G}_{t+1}$ into
a new context $C_{t+1}$:
\begin{equation}
C_{t+1} = \psi(\mathcal{G}_{t+1}),
\end{equation}
where $\psi(\cdot)$ maps active thoughts in $\mathcal{G}_{t+1}$ to their
corresponding contextual representations.
For folded trajectories, all associated reasoning steps in $C_t$ are replaced by
their summary thought $\bar{v}$.
For flushed nodes, verbose execution traces are substituted with compact thoughts
$\hat{v}$.
The resulting context is substantially smaller than $C_t$, while preserving the
logical structure required for coherent and goal-directed reasoning.

\begin{table*}[t]
    \centering
    \small
    \setlength{\tabcolsep}{4.5pt}
    \renewcommand{\arraystretch}{1.15}
   
    \begin{tabularx}{\textwidth}{l *{4}{>{\centering\arraybackslash}X} *{4}{>{\centering\arraybackslash}X}}
        \toprule
        \multirow{2}{*}{\textbf{Method}} &
        \multicolumn{4}{c}{\textbf{General AI Assistant}} &
        \multicolumn{4}{c}{\textbf{WebWalkerQA}} \\
        \cmidrule(lr){2-5}\cmidrule(lr){6-9}
        & \textbf{L1} & \textbf{L2} & \textbf{L3} & \textbf{Avg.} & \textbf{Easy} & \textbf{Med.} & \textbf{Hard} & \textbf{Avg.} \\
        \midrule

        \rowcolor{black!3}
        \multicolumn{9}{l}{\textbf{Direct Reasoning (w/o Retrieval)}} \\

        QwQ-32B                & 25.6 & 9.6  & 16.7 & 16.5 & 7.5  & 2.1  & 3.8  & 4.0 \\
        GPT-4o                 & 23.1 & 15.4 & 8.3  & 17.5 & 6.7  & 6.0  & 4.2  & 5.5 \\

        DeepSeek-R1-671B       & 43.6 & 26.9 & 8.3  & 31.1 & 5.0  & 11.8 & 11.3 & 10.0 \\

        \addlinespace[2pt]
        \rowcolor{black!3}
        \multicolumn{9}{l}{\textbf{Retrieval-Augmented Generation}} \\
        Vanilla RAG (QwQ-32B)  & 33.3 & 36.5 & 8.3  & 32.0 & 36.9 & 26.1 & 33.5 & 31.2 \\
        Query Planning (QwQ-32B)& 48.7 & 25.0 & 8.3  & 32.0 & 28.8 & 35.7 & 30.8 & 32.5 \\
        Iterative RAG (QwQ-32B)& 51.3 & 28.8 & 8.3  & 35.0 & 29.4 & 32.9 & 31.3 & 31.5 \\

        \addlinespace[2pt]
        \rowcolor{black!3}
        \multicolumn{9}{l}{\textbf{Tool-Integrated Reasoning}} \\
        ReAct (QwQ-32B)        & 48.7 & 34.6 & 16.7 & 37.8 & 35.6 & 29.1 & 13.2 & 24.1 \\
        ReAct (GPT-4o)         & 51.2 & 34.6 & 8.3  & 34.6 & 34.6 & 42.0 & 23.9 & 33.8 \\
        ReAct (Qwen3-30B-A3B)  & 48.7 & 26.9 & 8.3  & 33.0 & 26.3 & 27.5 & 21.7 & 25.2 \\
        WebThinker-32B~$\dagger$         & 56.4 & 50.0 & 16.7 & 48.5 & 58.8 & 44.6 & 40.4 & 46.5 \\
        WebDancer (QwQ-32B)~$\dagger$    & 56.4 & 48.1 & 25.0 & 46.6 & 49.4 & 55.0 & 29.6 & 43.2 \\
        ReSum-GRPO$~\dagger$            & --   & --   & --   & 48.5 & --   & --   & --   & -- \\
        DeepAgent-RL$~\dagger$         & 66.7 & 59.6 & 25.0 & 58.3 & --   & --   & --   & -- \\
        AgentFold-30B-A3B$~\dagger$      & --   & --   & --   & 67.0 & --   & --   & --   & -- \\
        GLM-4.6       & 76.9 & 59.6 & 33.3 & 63.1 & 64.4 & 62.9 & 48.8 & 58.2 \\
        DeepResearch-30B-A3B   & \underline{79.5} & 67.3 & 41.7 & 68.9 & \underline{72.5} & \underline{71.8} & \underline{61.3} & \underline{68.2} \\

\rowcolor{black!6}
\multicolumn{9}{l}{\textbf{MemoBrain-8B}} \\

\quad w/ GLM-4.6 &\underline{79.5} & \underline{71.2} & \underline{50.0} &  \underline{71.8} & 68.8 & 69.6 & \underline{61.3} & 66.5 \\

\quad w/ DeepResearch-30B-A3B &
\textbf{82.1} & \textbf{69.2} & \textbf{58.3} & \textbf{74.5} & \textbf{73.1} & \textbf{72.1} & \textbf{64.2} & \textbf{69.6} \\
        \bottomrule
    \end{tabularx}
     \caption{Main experimental results. Best scores are shown in \textbf{bold}, and second-best are \underline{underlined}. We report $\text{Pass}@1$. Results marked with $\dagger$ are cited from the original papers. MemoBrain results are obtained by integrating MemoBrain-8B into the corresponding base agents (GLM-4.6 and DeepResearch-30B-A3B).}
    \label{tab:exp}
    \vspace{-10pt}
\end{table*}

\begin{table}[t]
\centering
\small
\begin{tabular}{l c c}
\toprule
\textbf{LLM} & \textbf{Acc} & \textbf{Search Calls} \\
\midrule
Sonnet 4             & 37.35 & 9.03  \\
Kimi-K2              & 35.42 & 11.22 \\
oss-120b-high         & 44.10 & 18.35 \\
GLM-4.6              & 48.19 & 25.60 \\
DeepResearch-30B-A3B & 51.93 & 17.14 \\
\midrule
\textbf{MemoBrain} & & \\
\quad w/ GLM-4.6              & 55.06 & 24.62 \\
\quad w/ DeepResearch-30B-A3B & \textbf{60.36} & \textbf{27.12} \\

\bottomrule
\end{tabular}
\caption{Performance Comparison under Identical Retrieval Settings on BrowseComp-Plus}
\vspace{-10pt}
\label{tab:bcp}
\end{table}
\subsection{Optimization}
\label{sec:optimization}

MemoBrain is optimized with a two-stage training strategy that reflects the
distinct learning characteristics of memory construction and memory management.
Both functionalities are implemented by a single memory model $\phi$, which is
prompt-conditioned to operate under different roles.

\paragraph{Stage I: Memory Construction.}
Memory construction is treated as a high-frequency semantic abstraction problem.
Since this process mainly involves extracting salient outcomes from execution
traces and exhibits low decision ambiguity, we optimize $\phi$ using supervised
fine-tuning with high-quality thought annotations generated by a strong teacher
model (DeepSeek~V3.2):
\begin{equation}
\mathcal{L}_{\text{const}}
= - \mathbb{E}_{x_t}\,\log \phi(v_t \mid x_t).
\end{equation}
This stage equips MemoBrain with a stable and consistent abstraction capability,
which serves as the foundation for downstream memory control.

\paragraph{Stage II: Memory Management.}
Memory management is a sparse, decision-driven process that operates over the
trajectory memory $\mathcal{G}_t$ under a bounded context budget.
At each trigger point, MemoBrain predicts a set of memory operations
$\mathcal{O}_t$ (e.g., \textsc{Fold} or \textsc{Flush}) conditioned on the current
memory state. Because these decisions involve global trade-offs between compression and
semantic sufficiency, direct supervision is inadequate.

We therefore optimize memory management using Direct Preference Optimization~\citep{rafailov2024direct}.
For each memory state $\mathcal{G}_t$, we sample multiple candidate operation
sets, evaluate their downstream reasoning behavior, and construct preference
pairs $(\mathcal{O}^+, \mathcal{O}^-)$:
\begin{equation}
\mathcal{L}_{\text{mgmt}}
= - \mathbb{E}\,\log \sigma\!\left(
\log \frac{\phi(\mathcal{O}^+ \mid \mathcal{G}_t)}
{\phi(\mathcal{O}^- \mid \mathcal{G}_t)}
\right).
\end{equation}
% Together, this two-stage optimization strategy enables MemoBrain to learn
% reliable local abstraction while making effective executive decisions.

\section{Experiment}
\subsection{Datasets and Baselines}
\paragraph{Datasets.}
We evaluate MemoBrain and baslines on three benchmarks:
(1)~\textbf{GAIA}~\citep{mialon2023gaia} (General AI Assistant) consists of over 450 real-world queries covering multi-step reasoning, tool use, and multimodal understanding. 
Following prior work~\citep{agentfold,resum,deepagent}, we evaluate on 103 text-only validation questions. 
(2)~\textbf{WebWalker}~\citep{wu2025webwalker} comprises 680 queries across domains, requiring agents to traverse linked webpages and integrate evidence dispersed across multiple subpages. 

For GAIA and WebWalkerQA, Google Search is used as the \emph{online search tool}.

(3)~\textbf{BrowseComp-Plus}~\citep{bcp} is derived from the BrowseComp benchmark~\citep{wei2025browsecomp} and contains 830 questions whose answers are short and verifiable. 
For BrowseComp-Plus, a fixed candidate webpage corpus is provided, and agents perform local dense retrieval over this corpus as the search tool.

\paragraph{Baselines.}
We evaluate MemoBrain against representative baselines spanning three categories.
\emph{Direct Reasoning} includes strong standalone LLMs without external tools, such as QwQ-32B, GPT-4o, and DeepSeek-R1-671B~\citep{deepseekr1, qwen3,openai2024gpt4technicalreport}.
\emph{Retrieval-Augmented Generation} covers methods that inject retrieved evidence into the context, including vanilla RAG and enhanced variants with query planning or iterative refinement~\citep{ITER-RETGEN,chan2024rqraglearningrefinequeries}.
\emph{Tool-Integrated Reasoning} comprises approaches that interleave reasoning with retrieval actions, such as ReAct, Search-o1, and WebThinker~\citep{yao2022react,searcho1,li2025webthinker}, as well as more recent systems including WebDancer, ReSum, AgentFold, and DeepAgent~\citep{webdancer,resum,agentfold,deepagent}.  
Notably, ReSum, AgentFold, and DeepAgent introduce explicit forms of \emph{context management}, such as summarization, trajectory compression, or learned control, which represent the engineering realization of \emph{agent memory} in existing agent frameworks.

We further include GLM-4.6 and DeepResearch-30B-A3B as strong tool-augmented agent baselines~\citep{tongyi_ds,glm2024chatglm}.  
MemoBrain is integrated into these agents as an external memory component.
Implementation details are provided in Appendix~\ref{app:imp}.

\subsection{Main Results}
We summarize the main experimental results in Table~\ref{tab:exp} and Table~\ref{tab:bcp}, from which several consistent observations can be drawn.

(1)~\textbf{MemoBrain consistently improves the performance of its base agents} when integrated with both GLM-4.6 and DeepResearch-30B-A3B.
Across all evaluated benchmarks, MemoBrain yields clear gains over the original agent frameworks, demonstrating that introducing an explicit executive memory component is broadly beneficial for tool-augmented reasoning.
In particular, when combined with DeepResearch-30B-A3B, MemoBrain achieves the \emph{best overall performance across all benchmarks}, outperforming all baselines.
These results indicate that MemoBrain is adaptive to different reasoning agents and can be effectively incorporated as a plug-in memory module without architecture-specific customization.

(2)~MemoBrain yields \textbf{larger gains in harder reasoning regimes}, notably on BrowseComp-Plus and the Level-3 split of GAIA, where long-horizon reasoning and extensive tool interaction strain the raw context window of the reasoning agent. By explicitly modeling reasoning structure and providing just-in-time, task-relevant context, MemoBrain mitigates cognitive overload and enables deeper exploration. This is further evidenced in Table~\ref{tab:bcp}, where MemoBrain-augmented agents make more valid tools therefore achieving obvious improvement, indicating more effective and sustained tool use under bounded context constraints.

(3)~Compared to prior context management approaches such as DeepAgent and AgentFold, \textbf{MemoBrain is more effective by decoupling memory from the reasoning agent}. Existing methods embed folding or summarization directly into a single agent, entangling memory control with reasoning execution. In contrast, MemoBrain adopts a copilot-style design with an explicit executive memory that externally manages reasoning trajectories, yielding stronger empirical performance and a more flexible, reusable memory module across diverse tool-augmented agent frameworks.
\subsection{Ablation Study}
\label{sec:ablation}
\begin{figure}
    \centering
    \includegraphics[width=\linewidth]{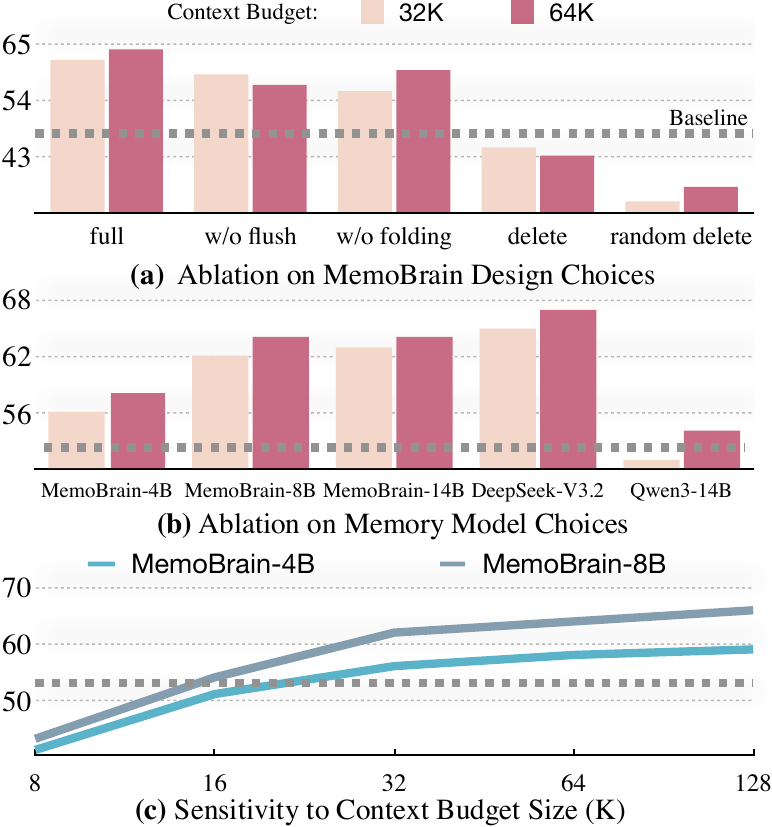}
\caption{Ablation studies of MemoBrain. All experiments are conducted on 100 randomly sampled instances from BrowseComp-Plus, using DeepResearch-30B-A3B as the base reasoning agent.}
\vspace{-5pt}
\label{fig:abl}
\end{figure}
To justify the technical validity of MemoBrain, we conduct comprehensive ablation studies along three dimensions:
\textbf{(a)} executive memory design choices,
\textbf{(b)} memory model selection, and
\textbf{(c)} sensitivity to memory budget size.
The results are in Figure~\ref{fig:abl}.

\paragraph{Impact of Executive Memory Design.}
We first ablate key design components of MemoBrain to assess their individual contributions.
Removing any major component consistently leads to performance degradation, indicating that MemoBrain’s effectiveness arises from their joint design.
Specifically, when retaining the memory graph structure and enabling only one operation, either \emph{trajectory folding} or \emph{selective flush}, performance improves over the baseline without executive memory.
In contrast, strategies such as directly deleting flushed sessions or randomly removing reasoning steps significantly undermine performance.
These highlight that MemoBrain’s gains stem not merely from reducing context length, but from \emph{structured and semantics-aware} memory operations that preserve the logical backbone of reasoning trajectories.
\begin{figure}
    \centering
    \includegraphics[width=\linewidth]{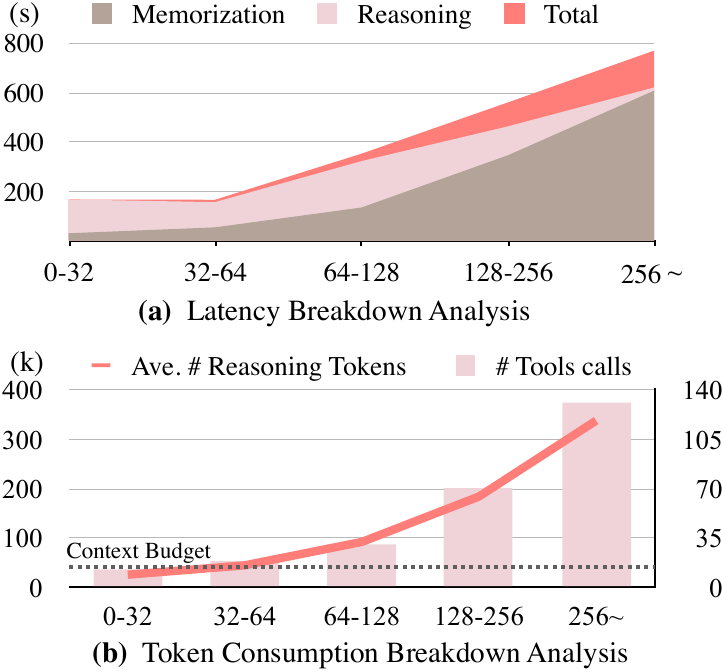}
    \caption{Efficiency Analysis of MemoBrain.
End-to-end runtime statistics on the full BrowseComp-Plus benchmark under a 32K memory budget, using 4×H100 GPUs for the reasoning model and 2×H100 GPUs for the memory model with a batch size of 8.}
    \label{fig:efficiency}
    \vspace{-10pt}
\end{figure}
\paragraph{Impact of Memory Model Choice.}
We evaluate MemoBrain instantiated with Qwen3 backbones of different scales (4B, 8B, and 14B).
In all cases, integrating MemoBrain leads to clear performance improvements over using the reasoning agent alone, demonstrating that the benefits of executive memory are not tied to a specific model scale.
While employing a stronger memory model such as DeepSeek-V3.2 achieves the best overall results, MemoBrain with 8B or 14B backbones exhibits comparable performance.
Considering this trade-off between efficiency and effectiveness, we adopt MemoBrain-8B in most experiments.

Moreover, we observe that models lacking task-specific optimization exhibit weaker instruction-following behavior (e.g., Qwen3-14B), often producing invalid memory updates or inappropriate management operations that disrupt reasoning trajectories.
This suggests that effective executive memory relies more on targeted optimization than on model scale alone.

\paragraph{Sensitivity to Memory Budget Size.}
Finally, we analyze MemoBrain’s sensitivity to the context budget.
Across all settings, larger budgets generally lead to better performance.
When the budget is set below 16K, performance can drop below the baseline, as the retained context becomes insufficient to support meaningful reasoning expansion.
Beyond 64K, however, performance gains begin to saturate, suggesting that the memory budget is sufficient for most tasks and that memory management may be triggered less frequently.
This trend indicates that MemoBrain is robust to budget selection within a reasonable range and does not require excessively large memory to be effective.

% Overall, these ablation results validate the technical soundness of MemoBrain’s design and confirm that its performance gains arise from principled executive memory modeling rather than superficial context reduction.

\subsection{Efficiency Analysis.}
We evaluate the efficiency of MemoBrain with respect to effective reasoning context length (Figure~\ref{fig:efficiency}).
Figure~\ref{fig:efficiency}~(a) presents a latency breakdown into reasoning time (reasoning agent runtime), memorization time (asynchronous memory construction), and total time, which additionally includes memory management when triggered.
Because memorization operates asynchronously, it incurs no end-to-end latency overhead as long as its cost remains below the reasoning time.
For trajectories up to 128K tokens, both memorization and management overheads are negligible compared to reasoning.
Although memory-related latency increases for longer trajectories as the memory graph grows, memorization remains consistently non-blocking even beyond 256K tokens, confirming that MemoBrain scales efficiently under a copilot-style, asynchronous execution regime.

Figure~\ref{fig:efficiency}~(b) reports token consumption and tool usage.
Although the base agent is limited to a 128K context window, integrating MemoBrain enables effective reasoning over hundreds of thousand tokens and supports substantially more tool calls, while operating under a smaller fixed context budget.
This result indicates that MemoBrain extends the practical reasoning horizon of tool-augmented agents by delivering compact, just-in-time contextual support that preserves essential reasoning structure, rather than relying on unbounded raw context expansion.

\section{Conclusion}

In this paper, we study \emph{complex reasoning} in tool-augmented agent frameworks, where long-horizon tool interactions lead to the accumulation of reasoning traces and transient artifacts that strain bounded LLM contexts and impair task alignment and reasoning coherence.
To address this challenge, we introduce \emph{executive memory}, which reconceptualizes memory as an explicit control mechanism over reasoning structure rather than passive context storage.
Building on this perspective, we propose \emph{MemoBrain}, a copilot-style memory system that maintains a global reasoning graph and applies budget-aware folding and flushing to provide compact, just-in-time context.
Experiments on multiple challenging benchmarks demonstrate consistent and substantial performance gains, while ablation and efficiency studies further validate the effectiveness and scalability of the proposed design.
Additional case studies are provided in Appendix~\ref{app:case}.
\clearpage
\section*{Limitations}
Despite the demonstrated effectiveness of MemoBrain across multiple benchmarks, several limitations remain that are important to acknowledge.

(1) MemoBrain assumes that the underlying reasoning agent is capable of sustained, long-horizon tool use and iterative reasoning. However, many current LLMs tend to terminate reasoning prematurely after a limited number of tool calls, often drawing conclusions before sufficient evidence is collected. In such cases, the accumulated context does not reach the threshold required to trigger memory management, and MemoBrain has little opportunity to intervene. We observe this limitation in experiments with both Qwen3-30B-A3B-2507-Instruct and Qwen3-30B-A3B-2507-Thinking, where early termination restricts the practical benefits of executive memory.

(2) While the memory graph maintained by MemoBrain naturally supports richer cognitive operations inspired by human memory, such as reactivating previously flushed reasoning episodes or partitioning the memory graph for parallel exploration, this work focuses on only two core operations: folding and selective flush. These design choices are intended to isolate the fundamental effects of executive memory control. Exploring more advanced memory operations and their coordination remains an important direction for future work.

(3) Our experimental comparison does not include all potentially relevant baselines. Some memory-augmented methods are designed mainly for long-context settings and cannot be readily adapted to dynamic reasoning trajectories. Others lack public implementations or require computational resources beyond our constraints. Additionally, our reproduced results for DeepResearch-30B-A3B on GAIA and WebWalker are slightly lower than those reported in the original paper, partly due to differences in tool implementations (e.g., search quality and code execution). Nevertheless, all comparative experiments are conducted under identical settings, ensuring that the relative improvements by MemoBrain remain valid and meaningful.

\bibliography{custom}

\appendix

\begin{table*}[t]
\centering
\begin{tcolorbox}[
  width=\linewidth,
  colback=gray!5,
  colframe=gray!60,
  boxrule=0.5pt,
  arc=3pt,
  left=6pt,
  right=6pt,
  top=6pt,
  bottom=6pt,
  fontupper=\small\ttfamily
]
You are a Memory Agent collaborating on a complex multi-step research task. Your job is to append the latest reasoning step to the ongoing reasoning graph by creating new node(s) and edge(s) as needed.

The reasoning graph is composed of:

- Nodes:

\quad  - "task": the original user question or main task.
  
  \quad- "subtask": a subproblem decomposed during the reasoning process.
  
  \quad- "evidence": a factual answer or conclusion obtained during reasoning to address a (sub)task.
  
  \quad- "summary": a summary node synthesizing a set of previous nodes. (NOTE: You will NOT create a new summary node in this step.)

- Edges:

  \quad- links a "task" node to newly created "subtask" nodes (breaking down a task).
  
  \quad- links a "task" or "subtask" node to a more precise/reformulated "subtask".
  
  \quad- links an "evidence" node to the task or subtask it supports or answers.

Input (You will be given):

1. The current state of the reasoning graph (in human-readable or JSON form), including all current nodes and their types and edges.

2. A new reasoning step (subtask, intermediate result, or evidence) that must be appended to the graph.

Your goals:

1. Decide what new node(s) (subtask or evidence) should be created from the current reasoning step.
   \quad - Clearly identify the node "kind": "subtask" or "evidence".
   
   \quad - Take notes to summarize the core information of the reasoning step in the form of a list like [{"role": "", "content": ""}, ...],
   
   \quad - Do NOT create "summary" type nodes in this step.
   
2. Decide what new edge(s) are needed to connect the new node(s) to existing nodes or to other new nodes:
\quad 
  \quad  - For each edge, specify the "src" and "dst" (src is the node id, dst is the tmp\_id), and optionally a short "rationale" for this edge.
   
3. Output your update as a single JSON object with the following fields:

  \quad  - "add\_edges": a list of new edge objects with at least "src", "dst", and optionally "rationale".

Rules:

\quad - Output must be strictly valid JSON; do NOT include any commentary or explanation.

\quad - You may use existing node ids as the src of edges as appropriate.

\quad - Avoid creating cycles in the graph.

\quad - Only create new nodes corresponding to meaningful new steps in the reasoning.
\end{tcolorbox}
\caption{System prompt used for executive memory construction in MemoBrain.}
\label{prompt:memorize}
\end{table*}
\section{Appendix}
\subsection{Implementation Details}
\label{app:imp}
MemoBrain is initialized from the Qwen3 instruction-tuned model family~\citep{qwen3}, including \href{https://huggingface.co/Qwen/Qwen3-4B-Instruct-2507}{Qwen3-4B-Instruct-2507}, \href{https://huggingface.co/Qwen/Qwen3-8B}{Qwen3-8B}, and \href{https://huggingface.co/Qwen/Qwen3-14B}{Qwen3-14B}. Training data are synthesized from the InfoSeek benchmark~\citep{xia2025opendatasynthesisdeep}, which contains over 50K complex reasoning QA pairs; during synthesis, we use DeepSeek-V3.2 as the memory oracle to generate executive supervision signals, resulting in 37,719 memory construction samples and 3,016 memory management samples. Data synthesis incurs an API cost of \$389. All models are trained using \href{https://github.com/hiyouga/LlamaFactory}{LlamaFactory} with a learning rate of $2\times10^{-5}$, gradient accumulation steps of 8, per-device batch size of 1, maximum gradient norm of 1.0, AdamW optimizer, 3 training epochs, and a maximum sequence length of 32K tokens. Memory construction is optimized via supervised fine-tuning, while memory management is optimized using preference-based learning.The total training time is 17, 21, and 27 hours for the 4B, 8B, and 14B models, respectively.

For GAIA and WebWalker, we employ the \href{https://developers.google.com/custom-search/}{Google Custom Search API} to retrieve the top-10 results and use \href{https://jina.ai/}{Jina.AI}’s web reader API to obtain webpage content, which is refined using \href{https://huggingface.co/Qwen/Qwen3-30B-A3B-Instruct-2507}{Qwen3-30B-A3B-Instruct-2507}. Code execution is handled via Python’s built-in \texttt{exec()} function. For BrowseComp-Plus, dense retrieval is performed using Qwen3-Embed-8B. 

In the main experiments, MemoBrain’s memory budget is set to 32K tokens for GAIA and WebWalker and 64K tokens for BrowseComp-Plus, triggering memory management once the accumulated context reaches the budget, while baseline methods retain their original context limits (e.g., 128K for DeepResearch-30B-A3B). We report Pass@1 as the evaluation metric and use GPT-4o-2024-08-06 as the LLM-based judge. All experiments are conducted on 8 NVIDIA H100 80G GPUs.
\subsection{Prompts}
In Table~\ref{prompt:memorize} and Table~\ref{tab:recall}, we present the memorization prompt and the memory management prompt used by MemoBrain, respectively. Additional prompts employed in this work, including the system prompt for complex reasoning and the evaluation prompt, are provided in our codes, available at \href{https://github.com/qhjqhj00/MemoBrain}{\textit{this repository}}.

\begin{table*}[t]
\centering
\begin{tcolorbox}[
  width=\linewidth,
  colback=gray!5,
  colframe=gray!60,
  boxrule=0.5pt,
  arc=3pt,
  left=6pt,
  right=6pt,
  top=6pt,
  bottom=6pt,
  fontupper=\small\ttfamily
]
Below is a memory graph for a complex task, detailing the key steps and their dependencies. The graph is presented in an indented structure, for example:

Node 1 [node type] [status] [node info]

\quad\textcolor{c1}{--[->]} Node 2 (an edge) [edge rationale (explaining the reasoning to Node 2)]

Each node includes its current status (active or inactive).

Your tasks:

\quad 1. For the current reasoning status, identify which nodes are redundant and can be flushed. Redundant nodes include: invalid reasoning steps (e.g., those that did not yield meaningful feedback), or nodes that have been superseded and refined by later steps. This operation is called "flush".

\quad 2. Identify which paths in the graph can be folded. If a subtask, after multiple reasoning steps, has arrived at a conclusive reasoning result (e.g., target information found, or evidence clearly unavailable as indicated by an evidence node), you may collapse the preceding related reasoning steps. This operation is called "folding".

Important constraint:

\quad - Do not include the same node in both flush\_ops and fold\_ops. Each node may only be the target of one operation, since both flush and fold operations result in the node becoming inactive.

{examples}

Guidelines:

\quad - Only consider nodes whose status is "active" for flush and fold operations; ignore other status nodes for these operations.

\quad - Do not assign the same node to more than one of flush\_ops or fold\_ops. You must ensure each active node appears in at most one operation.

\quad - For "flush\_ops", provide a concise rationale for why each node is redundant or superseded within the current reasoning context.

\quad - For "fold\_ops", group sequential related node ids that can be folded, and provide summary notes (as a list of dicts, each with a role and content).

\quad - Do NOT include any extra explanation or commentary outside the JSON.

\quad - Output only the JSON object, no additional formatting or text.
\end{tcolorbox}
\caption{System prompt used for executive memory management in MemoBrain.}
\label{tab:recall}
\end{table*}
\subsection{Case Study}
\label{app:case}
In Table~\ref{case:graph}, we present a representative memory graph composed of six reasoning episodes, with the root node corresponding to the original input task.
This case demonstrates how MemoBrain restructures a long and entangled reasoning trajectory into a set of semantically coherent episodes, while explicitly encoding their dependency relations within the memory graph.
Through executive memory construction and management, MemoBrain abstracts the evolving problem-solving process, including subtask decomposition, evidence validation, and hypothesis elimination, rather than retaining raw execution traces.
The resulting memory graph offers a cognitively interpretable representation of the reasoning process, preserving only high-salience conclusions and essential logical dependencies.
Notably, a trajectory that originally spans tens of thousands of tokens is compactly represented by only a few hundred tokens in the memory graph, highlighting MemoBrain’s ability to maintain reasoning coherence and task alignment under a bounded context budget.

In Table~\ref{case:append}, we further illustrate a concrete example of the \emph{memory construction} operation in MemoBrain.
Each reasoning episode, comprising internal deliberation, tool invocation, and tool responses, is abstracted into a single thought unit with a temporary identifier prior to integration into the global memory graph.
During this process, MemoBrain extracts the episode’s semantic outcome, assigns an appropriate thought type and activation state, and establishes explicit dependency relations with predecessor thoughts.
This construction procedure transforms transient traces into stable and interpretable memory units, forming the structural foundation for subsequent memory management and context regulation.
As shown in the case, memory construction enables MemoBrain to incrementally build a dependency-aware representation of the reasoning trajectory while discarding low-level details that are unnecessary for future reasoning.

In Table~\ref{case:management}, we present a concrete example of the \emph{memory management} operation applied to the constructed memory graph.
As the reasoning trajectory expands and the working context approaches the predefined budget, MemoBrain evaluates the global memory structure and performs executive operations to regulate information exposure.
Specifically, completed sub-trajectories whose associated subtasks have reached conclusive outcomes are folded into compact summary thoughts, while low-utility or superseded thoughts are selectively flushed to reduce contextual redundancy.
Crucially, these operations preserve the dependency structure among the remaining thoughts, ensuring that the global reasoning process remains coherent and aligned with the task objective.
By folding and flushing execution-level traces, MemoBrain condenses hundreds of tokens into a small set of high-salience memory units, enabling sustained long-horizon reasoning under a bounded context budget without sacrificing critical intermediate conclusions.
\begin{table*}[t]
\centering
\begin{tcolorbox}[
  width=\linewidth,
  colback=gray!5,
  colframe=gray!60,
  boxrule=0.5pt,
  arc=3pt,
  left=6pt,
  right=6pt,
  top=6pt,
  bottom=6pt,
  fontupper=\small\ttfamily
]
- Node 1: [task] \textcolor{c2}{[Active]} Begin to solve the task: Could you provide the name of the individual who:   

- As of December 2023, the individual was the coordinator of a research group founded in 2009.  

- Co-edited a book published in 2018 by Routledge.  

- The individual with whom they co-edited the book was a keynote speaker at a conference in 2019. 

- Served as the convenor of a panel before 2020. - Published an article in 2012. - Completed their PhD on the writings of an English writer. \textcolor{red}{[input task]}

    \textcolor{c1}{--[->]} Node 2 [Rationale: Break down the main task into a subtask to search for relevant information based on initial criteria.]

\quad- Node 2: [subtask] \textcolor{c2}{[Active]} [{'role': 'assistant', 'content': "Search for individuals co-editing a Routledge book in 2018, focusing on those with a PhD on an English writer's writings and other criteria."}, {'role': 'user', 'content': 'Return search 
results from queries about co-edited book 2018 Routledge, research group founded in 2009 coordinator 2023, and convenor panel 2019 keynote.'}]

\quad    \textcolor{c1}{--[->]} Node 3 [Rationale: The evidence from search results supports the subtask by providing potential leads or data to analyze further.]

\quad\quad- Node 3: [evidence] \textcolor{c2}{[Active]} [{'role': 'assistant', 'content': 'Analyze search results to identify potential candidates matching the criteria.'}, {'role': 'user', 'content': 'Found multiple results including Mathew Abbott (editor of Michael Fried and Philosophy: Modernism, Intention, and Theatricality, Routledge 2018), but need to check other details like research group coordinator, convenor of panel, article in 2012, and PhD on English writer.'}]

\quad    \textcolor{c1}{--[->]} Node 4 [Rationale: Break down the subtask into more precise search queries to find the individual matching all criteria.]

\quad\quad- Node 4: [subtask] \textcolor{c2}{[Active]} [{'role': 'assistant', 'content': 'Refine search queries to focus on specific criteria: co-edited a Routledge book in 2018, research group coordinator founded in 2009, and other details like convenor of panel befor
e 2020, article in 2012, PhD on English writer.'}, {'role': 'user', 'content': "Search for 'co-edited Routledge book 2018', 'research group founded 2009 coordinator', 'convenor panel 2019 keynote speaker', 'article published 2012', 'PhD thesis on English writer'."}]

\quad\quad    \textcolor{c1}{--[->]} Node 5 [Rationale: The evidence from search results supports the subtask by providing data to analyze for potential candidates matching the criteria.]

\quad\quad\quad- Node 5: [evidence] \textcolor{c2}{[Active]} [{'role': 'assistant', 'content': 'Search results from refined queries show potential matches, such as Mona Baker (co-coordinator of Genealogies of Knowledge Research Network founded in 2009) and Pooyan Ta
mimi Arab (co-edited Routledge Handbook of Material Religion in 2023, not 2018). Need to verify other criteria like convenor of panel before 2020, article in 2012, and PhD on English writer.'}, {'role': 'user', 'content': 'Results include various faculty profiles but none clearly match all criteria; further analysis required.'}]

\quad\quad    \textcolor{c1}{--[->]} Node 6 [Rationale: Break down the subtask into a more specific search query to find co-edited Routledge book in 2018 with keynote speaker in 2019.]

\quad\quad\quad- Node 6: [subtask] \textcolor{c2}{[Active]} [{'role': 'assistant', 'content': "Search for 'Routledge 2018 co-edited Keynote 2019' to find books published in 2018 by Routledge that are co-edited, with one co-editor giving a keynote at a conference in 2019."}, {'role': 'user', 'content': 'Return search results from query about Routledge 2018 co-edited book and keynote speaker in 2019.'}]

\quad\quad\quad    \textcolor{c1}{--[->]} Node 7 [Rationale: The evidence from search results supports the subtask by providing data to analyze for potential candidates matching the criteria.]

\quad\quad\quad\quad- Node 7: [evidence] \textcolor{c2}{[Active]} [{'role': 'assistant', 'content': 'Analyze search results from the query to identify potential matches for the criteria, such as co-edited Routledge book in 2018 and keynote speaker in 2019.'}, {'role': 'user', 'content': 'Search results show various faculty profiles and pages but no direct matches; need to refine further or check other sources.'}]

\textcolor{red}{[Memory Graph: 482 tokens; Original Reasoning Trajectory: 17,268 tokens.]}

\end{tcolorbox}
\caption{Memory graph example. Nodes correspond to memory thoughts produced by MemoBrain, each labeled with its thought type, activation state, and compressed representation. Directed edges (\textcolor{c1}{–[->]}) denote dependency relations that capture how reasoning steps build upon one another.}
\label{case:graph}
\end{table*}

\begin{table*}[t]
\centering
\begin{lstlisting}[
  language=json,
  basicstyle=\ttfamily\footnotesize,
  breaklines=true,
  frame=single,
]
{
  "add_nodes": [
    {
      "tmp_id": "tmp6",
      "kind": "subtask",
      "thought": [
        {
          "role": "assistant",
          "content": "Search for 'Routledge 2018 co-edited Keynote 2019' to find books published in 2018 by Routledge that are co-edited, with one co-editor giving a keynote at a conference in 2019."
        },
        {
          "role": "user",
          "content": "Return search results from query about Routledge 2018 co-edited book and keynote speaker in 2019."
        }
      ]
    },
    {
      "tmp_id": "tmp7",
      "kind": "evidence",
      "thought": [
        {
          "role": "assistant",
          "content": "Analyze search results from the query to identify potential matches for the criteria, such as a co-edited Routledge book in 2018 and a keynote speaker in 2019."
        },
        {
          "role": "user",
          "content": "Search results show various faculty profiles and pages but no direct matches; further refinement or alternative sources are required."
        }
      ]
    }
  ],
  "add_edges": [
    {
      "src": 4,
      "dst": "tmp6",
      "rationale": "Decompose the original task into a focused search for a Routledge co-edited book published in 2018 with a keynote speaker in 2019."
    },
    {
      "src": "tmp6",
      "dst": "tmp7",
      "rationale": "The retrieved search results provide evidence to evaluate whether the subtask criteria are satisfied."
    }
  ]
}
\end{lstlisting}

\caption{Illustration of memory construction. Newly generated thoughts are first assigned temporary IDs and later reindexed upon insertion into the memory graph. Directed edges represent dependency relations between thoughts.}
\label{case:append}
\end{table*}

\begin{table*}
    \begin{lstlisting}[
  basicstyle=\ttfamily\footnotesize,
  breaklines=true,
  frame=single,
  label={lst:memory_ops}
]
{
  "flush_ops": [
    {
      "id": 2,
      "rationale": "This node is superseded by Node 20, which provides conclusive evidence that Laura Lojo-Rodriguez is the coordinator of the Discourse and Identity (D&I) research group, matching the first clue. Node 2's subtask has been successfully addressed."
    },
    {
      "id": 3,
      "rationale": "This node is superseded by Node 30 and subsequent nodes (31-52), which provide a comprehensive summary and detailed evidence regarding the co-edited book clue. Node 3's subtask has been refined and addressed through more specific searches."
    },
    {
      "id": 47,
      "rationale": "This node is redundant because its subtask to verify the 2012 article publication has been addressed by evidence nodes (48-50), which consistently show no direct evidence of a 2012 article, making further exploration unnecessary."
    }
  ],
  "fold_ops": [
    {
      "ids": [48, 49, 50],
      "rationale": "These nodes provide conclusive evidence regarding the 2012 article publication subtask. All searches consistently indicate no confirmed publication in 2012.",
      "notes": [
       {
          "role": "assistant",
          "content": "The objective was to determine whether Laura Lojo-Rodriguez published an academic article in 2012. The investigation involved targeted name-year search queries and a systematic review of her publicly available publication records."
        },
        {
          "role": "user",
          "content": "No evidence of a 2012 publication was identified. While publications from other years are documented, the 2012 criterion is conclusively ruled out based on consistent negative findings across all sources."
        }
      ]
    }
  ]
}
\end{lstlisting}

\caption{Illustration of memory management. MemoBrain applies executive operations to the memory graph by flushing low-utility thoughts and folding completed sub-trajectories into compact summary nodes. }
\label{case:management}
\end{table*}

\end{document}